\documentclass[sigconf]{acmart}
\usepackage{algorithm}
\usepackage{algpseudocode}
\usepackage{multirow}
\usepackage{multicol}
\AtBeginDocument{%
  }

\copyrightyear{2025}
\acmYear{2025}
\setcopyright{acmlicensed}
\acmConference[MM '25] {Proceedings of the 33nd ACM International Conference on Multimedia}{October 27--October 31, 2025}{Dublin, lreland.}
\acmBooktitle{Proceedings of the 33nd ACM International Conference on Multimedia (MM '25), October 27--October 31, 2025, Dublin, lreland}
\acmDOI{10.1145/XXXXXX.XXXXXX}

\setcopyright{acmlicensed}
\copyrightyear{2025}
\acmYear{2025}
\acmDOI{XXXXXXX.XXXXXXX}

\begin{document}

\title{A Unified Framework for Stealthy Adversarial Generation via  Latent Optimization and Transferability Enhancement}

\author{Gaozheng Pei}
\email{peigaozheng23@mails.ucas.ac.cn}
\affiliation{%
  \institution{EECE, UCAS}
  % \city{Beijing}
  \state{Beijing}
  \country{China}
}
\author{Ke Ma}
\email{make@ucas.ac.cn}
\authornotemark[1]
\affiliation{%
  \institution{EECE, UCAS}
  \state{Beijing}
  \country{China}
}
\author{Dongpeng Zhang}
\email{zhangdongpeng24@mails.ucas.ac.cn}
\affiliation{%
  \institution{EECE, UCAS}
  \state{Beijing}
  \country{China}
}
\author{Chengzhi Sun}
\email{sunchengzhi24@mails.ucas.ac.cn}
\affiliation{%
  \institution{EECE, UCAS}
  \state{Beijing}
  \country{China}
}
\author{Qianqian Xu}
\email{ xuqianqian@ict.ac.cn}
\affiliation{%
  \institution{IIP, ICT, CAS}
  \state{Beijing}
  \country{China}
}
\author{Qingming Huang}
\authornotemark[1]
\email{ qmhuang@ucas.ac.cn}
\affiliation{%
  \institution{ SCST, UCAS\\
 IIP, ICT, CAS\\
 BDKM, CAS}
  \state{Beijing}
  \country{China}
}

\renewcommand{\shortauthors}{Trovato et al.}

\begin{abstract}
Due to their powerful image generation capabilities, diffusion-based adversarial example generation methods through image editing are rapidly gaining popularity. However, due to reliance on the discriminative capability of the diffusion model, these diffusion-based methods often struggle to generalize beyond conventional image classification tasks, such as in Deepfake detection.  Moreover, traditional strategies for enhancing adversarial example transferability are challenging to adapt to these methods.  To address these challenges, we propose a unified framework that seamlessly incorporates traditional transferability enhancement strategies into diffusion model-based adversarial example generation via image editing, enabling their application across a wider range of downstream tasks. Our method won first place in the "1st Adversarial Attacks on Deepfake Detectors: A Challenge in the Era of AI-Generated Media" competition at ACM MM25, which validates the effectiveness of our approach.
\end{abstract}

\begin{CCSXML}
<ccs2012>
   <concept>
       <concept_id>10010147.10010178.10010224.10010245</concept_id>
       <concept_desc>Computing methodologies~Computer vision problems</concept_desc>
       <concept_significance>500</concept_significance>
       </concept>
 </ccs2012>
\end{CCSXML}

\ccsdesc[500]{Computing methodologies~Computer vision problems}

\keywords{Diffusion Model; Adversarial Examples; Deepfake Detection}


\maketitle
\renewcommand{\thefootnote}{\fnsymbol{footnote}}
\footnotetext[1]{Corresponding authors.}
\section{Introduction}
Deep neural networks (DNNs), despite their superior performance across diverse tasks \cite{zou2023object,mo2022review,prabhavalkar2023end,wang2024driving}, remain highly vulnerable to adversarial examples—inputs crafted by adding imperceptibly small perturbations that can cause models to make erroneous predictions \cite{Adversarial,szegedy2013intriguing,cw,PGD,autoattack}. Crucially, these adversarial perturbations frequently exhibit transferability: examples designed to mislead one model can often deceive other, potentially unknown models. This property, known as adversarial transferability \cite{dong2018boosting,distribution,xie2019improving,pei2025exploring,wuskip}, significantly heightens practical security concerns for deployed machine learning systems. It enables practical black-box attacks \cite{papernot2017practical,maho2021surfree}, where an attacker can compromise models without access to their internal parameters or architectures. Therefore, advancing the understanding of how to generate highly transferable adversarial examples is fundamental to developing more robust DNNs. \\
Traditional methods for enhancing the transferability of adversarial examples typically operate in the pixel space, and these approaches can be broadly categorized into five types. Gradient-based methods (e.g., MI-FGSM \cite{mifgsm}, VMI-FGSM \cite{Variancefgsm}, NEF \cite{mef}) optimize perturbations through momentum and variance tuning, while input transformation techniques (e.g., DIM \cite{xie2019improving}, SIM \cite{SIM}, TIM \cite{TIM}) apply stochastic augmentations to improve generalization. Advanced objective-based attacks \cite{tap,ILPD} exploit intermediate-layer features, neuron importance, or feature-space discrepancies to craft perturbations. Model-Specific attacks \cite{SGM,distribution} leverage architectural characteristics of surrogate models to amplify gradient propagation or selectively mask model parameters. Ensemble-based attacks \cite{Ens,Ghost} aggregate gradients or predictions from multiple surrogate models, which reduce bias toward any single model. However, the adversarial samples generated by the aforementioned methods exhibit limited stealthiness, making them perceptible to human observers.\\
Diffusion models \cite{ho2020denoising,rombach2022high,esser2024scaling}, owing to their powerful image generation capabilities, have been widely applied to numerous downstream tasks. Recent works have leveraged diffusion-based image editing techniques to construct adversarial examples \cite{chen2024diffusion,xue2023diffusion,zhou2024stealthdiffusion,dai2024advdiff,li2024advad}, thereby enhancing their stealthiness. However, these methods' frameworks are inherently incompatible with traditional adversarial transferability enhancement strategies and struggle to generalize beyond traditional classification tasks, significantly limiting their practical attack effectiveness in real-world scenarios.
To address these limitations, we propose a universal framework that is not only compatible with conventional adversarial transferability enhancement strategies but also applicable to arbitrary classification tasks. Specifically, to ensure that the edited image is as similar as possible to the original image, we first use DDIM inversion \cite{ddim-inversion} to obtain the latent variables. We select the latent at a certain timestep as the update target while freezing the latents of previous timesteps, using the image description provided by the vision-language model as textual guidance during the image restoration process. When optimizing the selected latent, suitable adversarial transferability approaches can be randomly assembled from pools of traditional methods. Overall, our framework offers two key advantages:
\begin{itemize}
    \item \textbf{Universal Compatibility}\quad It seamlessly integrates traditional and enhanced adversarial transferability methods.
    \item \textbf{Broad Applicability}\quad  Our framework generalizes to all kinds of downstream classification tasks.
\end{itemize}

\section{Related Work}
\subsection{Traditional Methods}
\textbf{Gradient Optimization-based Attacks} \quad These methods \cite{mifgsm,Variancefgsm,mef} improve adversarial transferability by optimizing gradient computation through techniques that stabilize update directions and mitigate local optima convergence.\\
\textbf{Input Transformation-based Attacks}\quad This category \cite{xie2019improving,SIM,TIM} improves transferability by diversifying input patterns via stochastic transformations, making adversarial examples less dependent on specific model architectures.\\
\textbf{Advanced Objective-based Attacks}\quad These approaches \cite{tap,ILPD} exploit intermediate-layer features, neuron importance, or feature-space discrepancies to craft perturbations that are more transferable across models.\\
\textbf{Model-specific Attacks} \quad These methods \cite{SGM,distribution} leverage the architectural characteristics of surrogate models to amplify gradient propagation or selectively mask model parameters.\\
\textbf{Ensemble-based Attacks} \quad By aggregating gradients or predictions from multiple surrogate models, ensemble methods \cite{Ens,Ghost} reduce bias toward any single model.
\subsection{Diffusion-based Methods}
While traditional methods have demonstrated significant improvements in enhancing the transferability of adversarial examples, their dependence on pixel-level modifications results in generated samples with limited stealthiness, rendering them perceptible to human observers. \cite{chen2024diffusion} proposes DiffAttack, the first method to leverage diffusion models for adversarial attacks by crafting perturbations in the latent space rather than pixel space, achieving both superior imperceptibility and enhanced transferability across black-box models. AdvDiff \cite{dai2024advdiff} introduces a novel approach to generating unrestricted adversarial examples by leveraging diffusion models' reverse generation process, proposing two interpretable adversarial guidance techniques that inject target-class gradients while preserving generation quality. Diff-PGD \cite{xue2023diffusion} introduces a diffusion-guided adversarial attack framework that leverages SDEdit purification during gradient optimization, enabling the generation of adversarial samples that maintain natural image distributions while achieving high attack success rates. StealthDiffusion \cite{zhou2024stealthdiffusion} introduces a latent-space adversarial optimization framework that combines Latent Adversarial Optimization (LAO) with a Control-VAE module to preserve image consistency.
\section{Preliminary}

% Text-guided diffusion models aim to map a random noise vector $z_t$ and textual condition $\mathcal{P}$ to an output image $z_0$, which corresponds to the given conditioning prompt.
% In order to perform sequential denoising, the network $\epsilon\theta$ is trained to predict artificial noise, following the objective: \\[-8pt]
% \begin{equation}
% \min_\theta E_{z_0,\epsilon\sim N(0,I),t\sim \text{Uniform}(1,T)} ||\epsilon-\epsilon_\theta(z_t,t,\mathcal{C})||^2_2.
% \end{equation}
% \\[-12pt]
% Note that $\mathcal{C} = \psi(\mathcal{P})$ is the embedding of the text condition and $z_t$ is a noised sample, where noise is added to the sampled data $z_0$ according to timestamp $t$.
% At inference, given a noise vector $z_T$, The noise is gradually removed by sequentially predicting it using our trained network for $T$ steps. 

Since we aim to accurately reconstruct a given real image, we employ the deterministic DDIM sampling \cite{songdenoising}: 
\begin{equation}
\mathbf{z}_{t-1} = \sqrt{\frac{\alpha_{t-1}}{\alpha_t}}\mathbf{z_t} + \left(\sqrt{\frac{1}{\alpha_{t-1}}-1}-\sqrt{\frac{1}{\alpha_t}-1}\right) \cdot \epsilon_\theta(\mathbf{z_t},t,\mathbf{\mathcal{C}}).
\end{equation}
Diffusion models often operate in the image pixel space where $\mathbf{z_0}$ is a sample of a real image. In our case, we use the popular and publicly available stable diffusion model \cite{rombach2022high} where the diffusion forward process is applied on a latent image encoding $\mathbf{z_0} = E(\mathbf{x_0})$ and an image decoder is employed at the end of the diffusion backward process $\mathbf{x_0} = D(\mathbf{z_0})$. \\
\textbf{Classifier-free guidance.}
One of the key challenges in text-guided generation is the amplification of the effect induced by the conditioned text. To this end, \cite{ho2021classifier} have presented the classifier-free guidance technique, where the prediction is also performed unconditionally, which is then extrapolated with the conditioned prediction. More formally, let $\mathbf{\varnothing} = \psi(``")$ be the \textit{embedding} of a null text and let $w$ be the guidance scale parameter, then the classifier-free guidance prediction is defined by: 
\begin{equation}
\tilde{\epsilon}_\theta(\mathbf{z_t},t,\mathbf{\mathcal{C}},\mathbf{\varnothing}) = w \cdot \epsilon_\theta(\mathbf{z_t},t,\mathbf{\mathcal{C}}) + (1-w) \ \cdot \epsilon_\theta(\mathbf{z_t},t,\mathbf{\varnothing}).
\end{equation}
\textbf{DDIM inversion.}
A simple inversion technique was suggested for the DDIM sampling \cite{dhariwal2021diffusion,songdenoising}, based on the assumption that the ODE process can be reversed in the limit of small steps: 
\begin{equation}
\mathbf{z_{t+1}} = \sqrt{\frac{\alpha_{t+1}}{\alpha_t}}\mathbf{z_t} + \left(\sqrt{\frac{1}{\alpha_{t+1}}-1}-\sqrt{\frac{1}{\alpha_t}-1}\right)\cdot \epsilon_\theta(\mathbf{z_t},t,\mathbf{\mathcal{C}}).
\end{equation}
In other words, the diffusion process is performed in the reverse direction, that is $\mathbf{z_0} \rightarrow \mathbf{z_T}$ instead of $\mathbf{z_T} \rightarrow \mathbf{z_0}$, where $\mathbf{z_0}$ is set to be the encoding of the given real image.  
\begin{figure*}[htbp]
    \centering
    \includegraphics[width=0.9\textwidth]{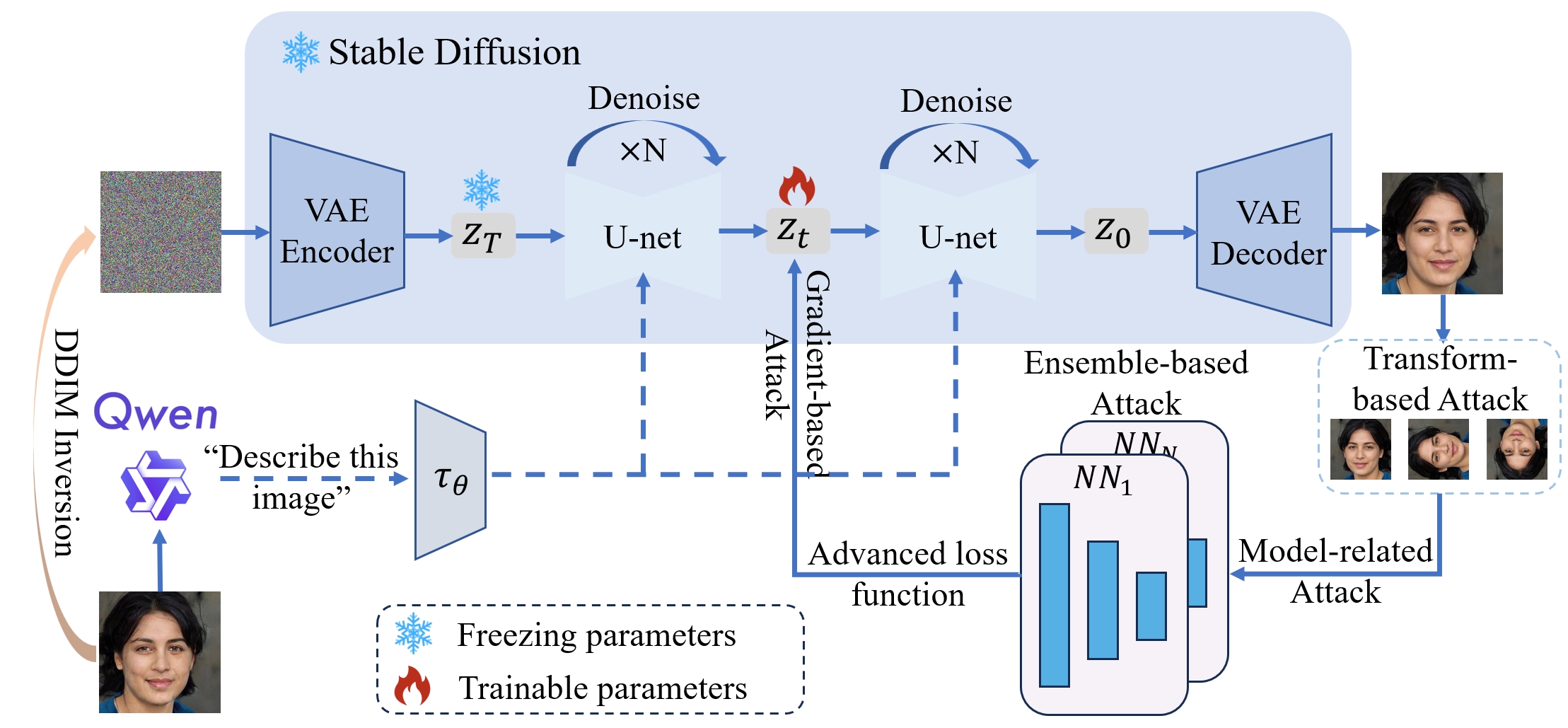}
    \caption{Framework of our method. The core of our framework lies in optimizing the latent representations at intermediate time-step to construct adversarial examples, while maintaining compatibility with conventional strategies for enhancing transferability.}
    \label{method}
\end{figure*}
\section{Methodology}
\subsection{Problem Formulation}
The objective of adversarial attacks against deepfake detectors is to craft adversarial perturbations that modify deepfake images—making them undetectable as synthetic content by state-of-the-art detectors—while maintaining high perceptual similarity to the original images. Given  deepfake images $\mathbf{x}$, we hope that we can craft adversarial perturbation $\delta$ to produce adversarial examples $\mathbf{x_{adv}}=\mathbf{x}+\delta$ such that the deepfake detector $f$ misclassifies $\mathbf{x_{adv}}$ as real label and the perturbation $\delta$ is imperceptible, maintaining high visual similarity between $\mathbf{x_{adv}}$ and the original $\mathbf{x}$. This problem can be formulated as follows: 
\begin{equation}
\begin{aligned}
    &\min \quad SSIM(\mathbf{x_{adv}},\mathbf{x})\\
    &s.t. \quad f(\mathbf{x_{adv}})=y_{real}.
\end{aligned}
\end{equation}
\subsection{Our Framework}
We display in the Figure \ref{method} the whole framework of our method, where we adopt the open-source stable diffusion \cite{rombach2022high} that pretrained on extremely massive text-image pairs. Since adversarial attacks aim to fool the target model by perturbing the initial image, they can be approximated as a special kind of real image editing \cite{diffedit}. Inspired by recent diffusion editing approaches. our framework leverages the DDIM Inversion technology \cite{ddim-inversion}, where the clean image is mapped back into the diffusion latent space by reversing the deterministic sampling process:
\begin{equation}
\label{inversion}
\begin{aligned}
    \mathbf{z_{t}} &= \operatorname{Inverse}(\mathbf{z_{t - 1}}) = \underbrace{\operatorname{Inverse} \circ \cdots \circ \operatorname{Inverse}}_{t}(\mathbf{z_{0}}),\\
    \mathbf{z_0} &= VAE-Encoder(\mathbf{x_0}),
\end{aligned}
\end{equation}
where $\operatorname{Inverse}$(·) denotes the DDIM Inversion operation. We apply the inversion for several timesteps from $\mathbf{x_0}$ (the initial image) to $\mathbf{x_t}$. A high-quality reconstruction of $\mathbf{x_0}$ can then be expected if we conduct the deterministic denoising process from $\mathbf{x_t}$. Different from  existing image editing approaches proposed to modify text embeddings for image editing, we choose to directly modify the latent at smaller timesteps, which achieves image editing while preserving the original semantic information of the image as much as possible. We here directly perturb the latent as follows: 
\begin{equation}
    \underset{\mathbf{z_{t}}}{\arg\min}\mathcal{L}_{\text{attack}}:=J\left(\mathbf{x_{adv}}, y_{true},f\right).
\end{equation}
 We can get $\mathbf{x_{adv}}$ as follows:
\begin{equation}
\label{denoise}
    \mathbf{x_{adv}}=VAE-Decoder(\mathbf{z_{0}^{\prime}})=\underbrace{\text{Denoise} \circ \cdots \circ \text{Denoise}}_{t}\left(\mathbf{z_{t}},\mathcal{C}\right),
\end{equation}
where $J$(·) is the cross-entropy loss and Denoise(·) denotes the diffusion denoising process. 
In addition to classification loss, other losses $\mathcal{L}_{others}$ can also be enhanced to maintain image consistency (e.g., L1 loss) or to improve the transferability of adversarial examples. The final loss can be formulated as follows:
\begin{equation}
\label{total-loss}
    \mathcal{L} = \mathcal{L}_{attack} + \lambda \mathcal{L}_{others}.
\end{equation}
To further enhance image consistency, we incorporate text guidance by utilizing a pre-trained vision-language model to describe the image content. The prompt "Describe this image" is used, and the resulting textual description serves as text guidance during the reverse process.\\
The complete algorithm flow can be referred to in \ref{framework}. As we can see, by decoupling the inverse image restoration process from adversarial attacks to the greatest extent, this allows our framework is compatible with almost all traditional strategies for enhancing adversarial example transferability, while not relying on the diffusion model's own discriminative capability for images. This makes it applicable to arbitrary downstream classification tasks.
\begin{algorithm}
\caption{Our Framework}
\begin{algorithmic}[1] % [1] ensures that line numbers are displayed
\item[\textbf{Input:}] Synthetic face $\mathbf{x}$, Real face label $y_{true}$, Perturbation radius $\epsilon$, Classifier $[f_1,f_2,...,f_N]$, Step size $\alpha$, Qwen $Q$.
\item[\textbf{Output:}] Adversarial face $\mathbf{x_{adv}}$, Iterations $T$.
\State Text embedding $\mathcal{C} = \psi(Q(x,"Describe\ this\ image"))$.
\State $\mathbf{z_t^0}$=DDIM-Inversion($\mathbf{x}$) via eq \eqref{inversion}.
\State Random initialization perturbation $\delta$.
\For {$i=0$ to $T$}
    \State Get clean image $\mathbf{x_{adv}^i}$ via eq \eqref{denoise}.
    \State //\textcolor{gray}{Transform-based attack}
    \State Image augmentation $\mathbf{x_{aug}^i}=\mathcal{T}(\mathbf{x_{adv}^i})$.
    \State //\textcolor{gray}{Advanced loss function and model-related attack}
    \State Calculate loss $loss_n = \mathcal{L}(f_n,\mathbf{x_{aug}},y_{true})$ via eq \eqref{total-loss}.
    \State //\textcolor{gray}{Ensemble-based attack}
    \State Ensemble loss $loss = Ensemble(loss_1,loss_2,...,loss_N)$.
    \State //\textcolor{gray}{Gradient-based attack}
    \State Update $\delta^{(i+1)} = \delta^i - \alpha\times\frac{\partial loss}{\partial\delta^i}$.
    \State Update $\delta^{(i+1)} = Proj(\delta^{(i+1)},-\epsilon,\epsilon)$.
    \State Update $\mathbf{z_t^{i+1}}=\mathbf{z_t}+\delta^{(i+1)}$.
\EndFor
\end{algorithmic}
\label{framework}
\end{algorithm} 
\section{Experiments}
\subsection{Experimental Setting}
\textbf{Dataset}\quad
The AADD-2025 Challenge provides a comprehensive dataset designed to evaluate the robustness of deepfake detection systems against adversarial attacks. The entire dataset is divided into sixteen subsets, categorized along two axes: generation method (GAN-based or diffusion-based) and image resolution (high quality or low quality). Each category contains four distinct image generators.All images in the dataset are synthetic, generated by state-of-the-art GAN and diffusion models, and exhibit a broad range of visual styles and qualities. This diversity ensures a challenging environment for both detection and adversarial attack tasks.\\
\textbf{Evaluation Metric}\quad
To quantitatively assess the effectiveness and imperceptibility of adversarial perturbations, the evaluation metric jointly considers two components: structural similarity and attack success rate. Specifically, each adversarial image is required to preserve high visual similarity to its original counterpart, measured by the Structural Similarity Index (SSIM), while simultaneously being misclassified as ``real'' by multiple deepfake detection systems.
Let $I_k$ denote the $k$-th image in the deepfake test dataset, and $I_k^{ADV}$ its corresponding adversarial version. Let $C = \{C_1, C_2, ..., C_m\}$ represent the set of classifiers. The final score is defined as:
\[
\sum_{C_f \in C} \sum_{k=1}^{N} SSIM(I_k, I_k^{ADV}) \cdot \left[ C_f(I_k^{ADV}) = LABEL_{real} \right],
\]
\noindent where:
\begin{itemize}
    \item $N$ is the total number of test images.
    \item $SSIM(I_k, I_k^{ADV})$ measures the perceptual similarity between the original and adversarial images.
    \item $C_f(I_k^{ADV})$ is the prediction of classifier on adversarial image.
    \item $LABEL_{real}$ denotes the ground truth label for real images.
    \item $\left[ \cdot \right]$ is the indicator function, which returns 1 if the condition is true, and 0 otherwise.
\end{itemize}
A high final score indicates that the adversarial image successfully bypasses the detection system while remaining visually close to the original, thereby reflecting both attack success and imperceptibility. \\
\textbf{Implementation details} \quad
To minimize content distortion when the attack succeeds, we adopt an attack strategy with a gradually increasing perturbation radius. The initial attack radius is set to 0.05, the final radius to 0.3, and the step size between iterations to 0.02. For the transform-based attack, we selected the following data augmentations from the third-party library Kornia:
RandomVerticalFlip, RandomHorizontalFlip, CenterCrop, RandomRotation90, RandomChannelDropout. For ensemble-based
attack, we combined the classification losses from different white-box classifiers by summing them with equal weights. For the advanced loss function, we additionally incorporated an L1 loss as a constraint to preserve image content consistency. We used MI-FGSM as our gradient-based attack. For the guidance scale parameter, we set $w=1$ as suggested by \cite{ddim-inversion} for better editability and reconstruction. We set DDIM sampling steps is 20 and we optimized the latent representations for either the 1st or 2nd timestep.

\subsection{Experimental Result}
\begin{figure}[htbp]
    \centering
    \includegraphics[width=0.45\textwidth]{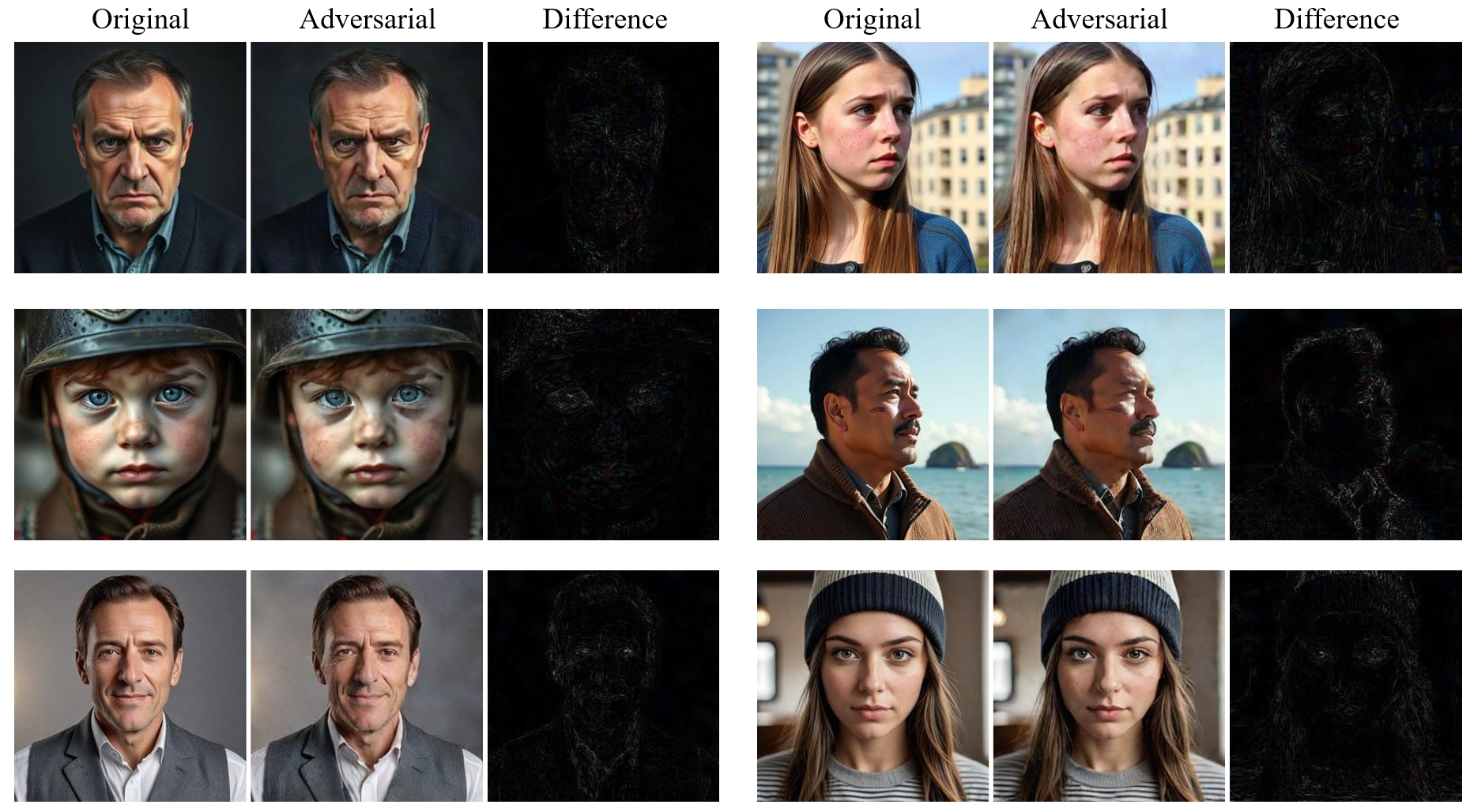}
    \caption{Visualization of the attack effectiveness of our method. }
    \label{visualization}
\end{figure}
\begin{table}[htbp]
  \centering
  \caption{Comparison of transfer attack success rate and SSIM between our method and the traditional method.}
    \begin{tabular}{cccc}
    \toprule
    \multirow{2}[0]{*}{Method} & \multicolumn{3}{c}{Resnet50->DenseNet121} \\
          & Radius & ASR ($\uparrow$) & SSIM ($\uparrow$) \\
          \midrule
    \multicolumn{1}{c}{\multirow{3}[0]{*}{Traditional}} & 4/255 & 5.23  & 0.8627 \\
          & 8/255 & 15.35 & 0.7876 \\
          & 16/255 & 43.38 & 0.6032 \\
    \midrule
    Ours  & -     & \textbf{100}   & 0.7814 \\
    \bottomrule
    \end{tabular}%
  \label{asr&ssim}%
\end{table}%
\begin{table}[htbp]
  \centering
  \caption{Comparison of final scores among the top 5 teams.}
    \begin{tabular}{lrcccc}
    \toprule
    Team name & \multicolumn{1}{c}{Ours} & \multicolumn{1}{l}
    {Safe AI} & \multicolumn{1}{l}{RoMa} & \multicolumn{1}{l}{GRADIANT}& \multicolumn{1}{l}{DASH}\\
    \midrule
    Final score ($\uparrow$) & \textbf{2740}  & 2709  & 2679 &2631&2618\\
    \bottomrule
    \end{tabular}%
  \label{final-score}%
\end{table}%
We first conducted a comparative analysis using two key metrics: SSIM and transfer success rate. As demonstrated in Table \ref{asr&ssim}, our method achieves a near-perfect transfer success rate of 100\%, while maintaining SSIM values comparable to traditional methods with a perturbation radius of 8/255. Notably, we observe that traditional methods fail to achieve successful transfer on certain samples even without attack radius constraints, which further validates the superiority of our method.
Visualization of attack effects (Figure \ref{visualization}) reveals that the perturbations generated by our method are primarily concentrated on foreground subjects while leaving background areas virtually unaffected, demonstrating the superior stealth characteristics of our adversarial examples.
The comprehensive evaluation in Table \ref{final-score} shows that our method achieved the highest final score among all 13 competing teams. These experimental results collectively confirm the effectiveness and robustness of our proposed approach.

\nocite{Pairs,liu2025futurepasttamingtemporal}
\bibliographystyle{ACM-Reference-Format}
\bibliography{sample-base}

\appendix

\end{document}